\documentclass[final]{cvpr}
\usepackage{caption}
\usepackage{times}
\usepackage{epsfig}
\usepackage{graphicx}
\usepackage{amsmath}
\usepackage{amssymb}
\usepackage{multirow} 
\usepackage{bm} 
\usepackage[misc]{ifsym}
\usepackage{url}

\usepackage[pagebackref=true,breaklinks=true,colorlinks,bookmarks=false]{hyperref}

\def\cvprPaperID{6705} 
\def\confYear{CVPR 2021}

\begin{document}

\title{Lifelong Person Re-Identification via Adaptive Knowledge Accumulation}
 
\author{Nan Pu\textsuperscript{1}~~~~Wei Chen\textsuperscript{1}~~~~Yu Liu\textsuperscript{2}\footnotemark[1]~~~~Erwin M. Bakker\textsuperscript{1}~~~~Michael S. Lew\textsuperscript{1}\\
\textsuperscript{1}LIACS Media Lab, Leiden University, The Netherlands\\\textsuperscript{2}International School of Information Science \& Engineering, Dalian University of Technology, China\\
{\tt\small \{n.pu, w.chen, erwin, m.s.k.lew\}@liacs.leidenuniv.nl, liuyu8824@dlut.edu.cn}
}
\maketitle

\begin{abstract}
Person re-identification (ReID) methods always learn through a stationary domain that is fixed by the choice of a given dataset. In many contexts (e.g., lifelong learning), those methods are ineffective because the domain is continually changing in which case incremental learning over multiple domains is required potentially. In this work we explore a new and challenging ReID task, namely lifelong person re-identification (LReID), which enables to learn continuously across multiple domains and even generalise on new and unseen domains. Following the cognitive processes in the human brain, we design an Adaptive Knowledge Accumulation (AKA) framework that is endowed with two crucial abilities: knowledge representation and knowledge operation. Our method alleviates catastrophic forgetting on seen domains and demonstrates the ability to generalize to unseen domains. Correspondingly, we also provide a new and large-scale benchmark for LReID. Extensive experiments demonstrate our method outperforms other competitors by a margin of 5.8\% mAP in generalising evaluation. The codes will be available at \url{https://github.com/TPCD/LifelongReID}.
\end{abstract}
\vspace{-2em}
\section{Introduction}
Person re-identification (ReID) seeks to linking the same pedestrian across disjoint camera views. While advanced deep learning methods \cite{zou2020joint, zhao2020unsupervised, porrello2020robust, yu2020weakly, wang2020unsupervised, pu2020dual, zhang2020relation} have shown powerful abilities for ReID~\cite{song2019generalizable, jin2020style}, their training process is limited heavily by a fixed and stationary 
dataset~\cite{zheng2015scalable,zheng2017unlabeled, wei2018person}. However, this limitation violates many practical scenarios where the data is continuously increasing from different domains. For instance, smart surveillance systems~\cite{zheng2016person, leng2019survey} over multiple crossroads capture millions of new images every day, and they are required to have the ability of incremental or lifelong learning.
\renewcommand{\thefootnote}{\fnsymbol{footnote}} 
\footnotetext[1]{Corresponding Author.} 

\begin{figure}[t]
  \begin{center}
  \includegraphics[width=232pt]{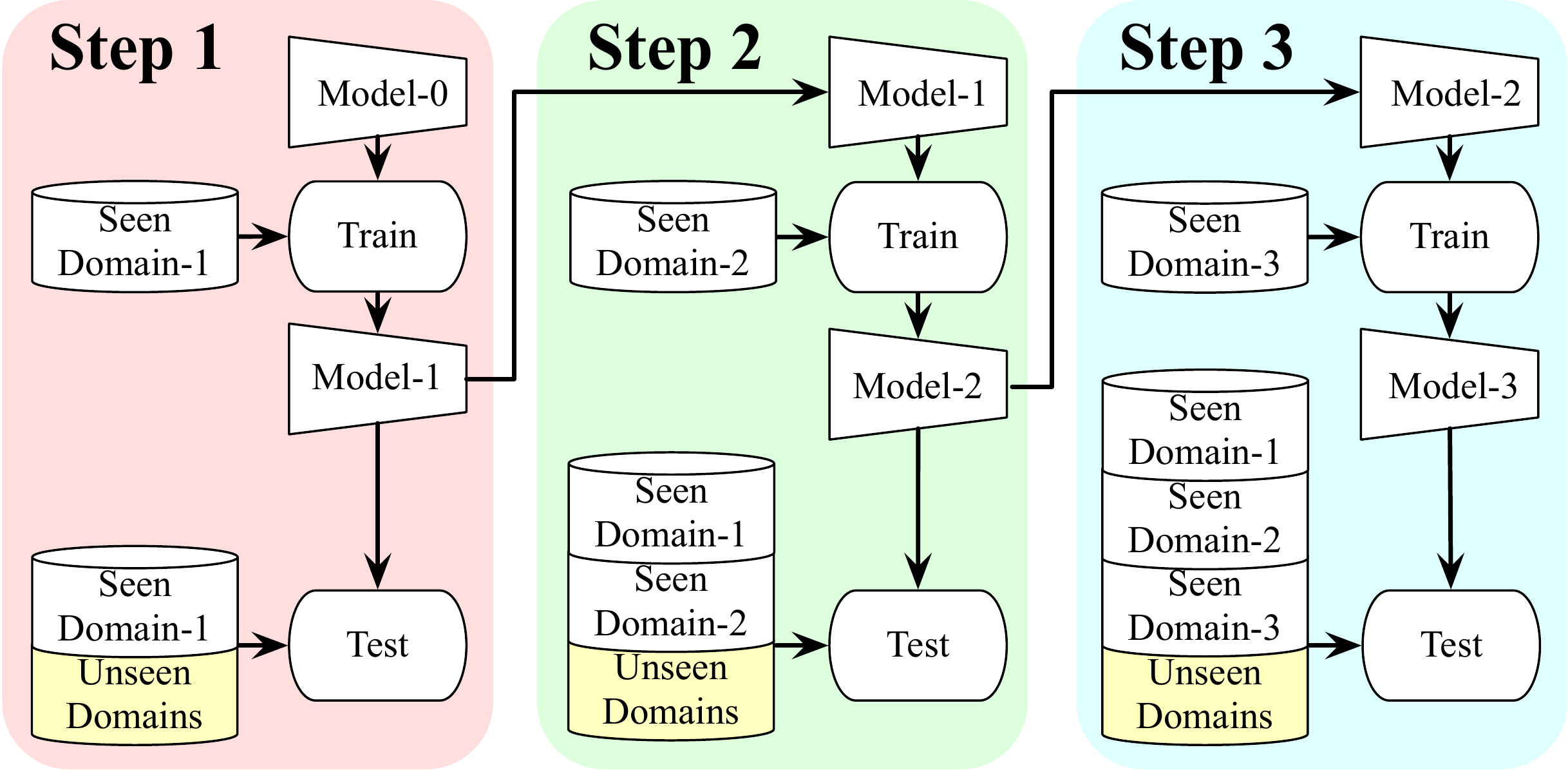}
  \captionsetup{aboveskip=0pt}
\caption{Pipeline of the proposed lifelong person re-identification task. The person identities among the involved domains are completely disjoint.
\label{fig:pipeline}}
  \end{center}
\vspace{-0.9cm}
\end{figure}

\par
To overcome the above limitation, we propose a new yet practical ReID task, namely \textit{lifelong person re-identification} (LReID), which requires the model to accumulate informative knowledge incrementally from several seen domains and then adapt the knowledge to the test sets of both seen and unseen domains (Fig.~\ref{fig:pipeline}). Our LReID task has two challenging problems, compared to previous tasks. First, unlike conventional lifelong learning \cite{mccloskey1989catastrophic,rebuffi2017icarl}, LReID further considers improving the generalization ability on unseen classes that never appear in the lifelong training stage. Second, LReID is a fine-grained lifelong learning task, in which inter-class appearance variations are significantly subtler than standard lifelong learning benchmarks like CIFAR-100 \cite{krizhevsky2009learning} and ImageNet \cite{russakovsky2015imagenet}.

\par
To tackle the challenges in LReLD, we propose a new \textit{adaptive knowledge accumulation} (AKA) framework which can continually accumulate knowledge information from old domains, so as to have a better generalization quality on any new domain. This idea is inspired by a new perspective of human cognitive processes. 
Recent discoveries \cite{cowell2019roadmap,wang2018knowledge} in cognitive science indicate that a cognitive process could be broadly decomposed into \textit{``representations''} and \textit{``operations''}. The structure of the knowledge representations (KRs) plays a key role for stabilizing memory, which shows our brain has potential relations with graph structure. Adaptive update and retrieval contained in the knowledge operations (KOs) promotes the efficient use of knowledge. Such complex yet elaborate KRs and KOs enable our brain to perform life-long learning well. 
Motivated by this, we endow AKA with two abilities to separately accomplish \textit{knowledge representation} and \textit{knowledge operation}. Specifically, we first represent transferable knowledge as a knowledge graph (KG), where each vertex represents one type of knowledge (e.g., the similar appearance between two persons). For image samples in one mini-batch, we temporally construct a similarity graph based on their relationships. Then, AKA establishes cross-graph links and executes a graph convolution. Such operation enables KG to transfer previous knowledge to each current sample. Meanwhile, KG is updated by summarizing the information underlying the relationships among current instances. Furthermore, for encouraging KG to improve learned representation while considering the forgetting problem, \textit{plasticity loss} and \textit{stability loss} are integrated to achieve an optimal balance for generalization on unseen domain. Our contributions 
are three-fold:
\par
\textbf{Task contribution.} We exploit a new yet practical person ReID task, namely LReID, which considers person re-identification problem under a lifelong learning scenario.
\par
\textbf{Technical contribution.} We propose a new AKA framework for LReID. AKA maintains a learnable knowledge graph to adaptively update previous knowledge, while transferring the knowledge to improve generalization on any unseen domains, with the plasticity-stability loss.
\par
\textbf{Empirical contribution.} We provide a new benchmark and evaluation protocols for LReID. AKA shows promising improvements over other state-of-the-art methods.
\section{Related Work}

\subsection{Person Re-identification Setups}
As summarized in Tab.~\ref{tab:lreid}, previous person ReID works are performed in four different setups: 1) Fully-supervised (FS) methods investigate and exploit different network structures 
and loss functions \cite{zheng2016person,pu2020dual,zhang2020relation,porrello2020robust}; 
2) Unsupervised domain adaption (UDA) is introduced to mitigate the domain gaps between source and target domain, caused by discrepancies of data distribution or image style~\cite{zheng2017unlabeled,wang2020unsupervised,zhao2020unsupervised,zou2020joint}; 3) Pure-unsupervised (PU) setting is less researched as it has to handle learning robust representation without using any label information~\cite{lin2019bottom}. 4) Domain generalization (DG) is an open-set problem. Lately, DG ReID task is explored by \cite{song2019generalizable}. 
However, all the above setups do not address the lifelong learning challenge in our LReID.
\par
The most related works \cite{li2019one} and \cite{zhao2020continual} proposed an online-learning method for one-pass person ReID and a continual representation learning setting for bio-metric identification, respectively. However, both of them focused on intra-domain continual learning instead of our inter-domain incremental learning. Since there are relatively narrow domain gaps between the training and the testing set, their settings are less challenging for keeping learned knowledge while improving generalization.

\begin{table}[th]
\small
\centering
\captionsetup{aboveskip=0pt}
\caption{The comparison of fully-supervised (FS), unsupervised domain adaption (UDA), pure unsupervised (PU), domain generalization (DG), and lifelong person re-identification (LReID). ``S.'' and ``T.'' denote source and target domain, respectively. \label{tab:lreid}}
\begin{tabular}{ccccc}
\hline
\multicolumn{1}{c}{Setup} & \multicolumn{1}{c}{Step}  & \multicolumn{1}{c}{Train} & \multicolumn{1}{c}{Label} & \multicolumn{1}{c}{Test} \\ \hline
FS \cite{zheng2016person} & one & S. & S.& S.\\
UDA \cite{wang2020unsupervised}                                & one or two    & S. \& T.   & S.      & T.       \\
PU \cite{lin2019bottom}   & one    & S.  & -     & S.      \\
DG \cite{song2019generalizable}                                & one           & all S.         & all S.      & T.       \\ \hline
LReID                                & multiple      & current S.     & current S.  & S. \& T. \\ \hline
\end{tabular}
\vspace{-2em}
\end{table}

\subsection{Lifelong Learning}
Lifelong or incremental learning \cite{pentina2014pac,castro2018end,parisi2019continual} dates back several decades, but now is attracting an ever-increasing attention due to impressive progresses in deep neural networks. Existing methods focus on common vision tasks like object recognition \cite{castro2018end,rebuffi2017icarl}, object detection \cite{shmelkov2017incremental} and image generation~\cite{wu2018memory}. The key challenge for lifelong learning is \textit{catastrophic forgetting}, which means that the model has performance degradation on previous tasks after training on new tasks. Existing methods can be divided into three categories, including knowledge distillation by the teacher-student 
structure \cite{li2017learning}, regularizing the parameter updates \cite{yoon2017lifelong} when new tasks arrive, and storing or generating 
image samples of previous tasks~\cite{rebuffi2017icarl,wu2018memory}.
\par
However, these methods are not suitable for LReID for various reasons. 1) The number of classes in ReID is much larger than that in conventional lifelong learning tasks, e.g., the popular benchmarks for them include MNIST~\cite{lecun1998gradient}, CORe50~\cite{lomonaco2017core50}, CIFAR-100~\cite{krizhevsky2009learning}, CUB~\cite{wah2011caltech} and ImageNet~\cite{russakovsky2015imagenet}. Except ImageNet, other benchmarks are small-scale in terms of classes numbers. In contrast, one of the popular ReID benchmarks, MSMT17\_V2 \cite{wei2018person} includes 4,101 classes/identities. 2) ReID datasets are more imbalanced~\cite{liu2020deep}, that means the number of samples per class ranges from 2 to 30. Because model degradation typically happens when learning from tail classes, LReID also suffers from a few-shot learning challenge. 3) Similar with the fine-grained retrieval task \cite{chen2020exploration}. The inter-class appearance variations for ReID are significantly subtler than generic classification tasks. It is particularly challenging in the lifelong learning scenario. 4) Previous works use the same classes for both training and testing, while ReID always need to handle with unseen classes. Fortunately, we find that remembering previously seen classes is beneficial for generalising on newly unseen classes.

\section{Lifelong Person Re-Identification}
\subsection{Problem Definition and Formulation}
In terms of LReID, one unified model needs to learn $T$ 
domains in an incremental fashion.
Suppose we have a stream of datasets $\mathcal{D} = \{D^{(t)}\}_{t=1}^{T}$. 
The dataset of the $t$-th domain is represented as $D^{(t)} = \{D^{(t)}_{tr}, D^{(t)}_{te}\}$, where $D^{(t)}_{tr}=\{ (\mathbf{x}_{i},\mathbf{y}_{i})\}_{i=1}^{\left|D^{(t)}_{tr}\right|}$ contains training images and their corresponding labels set $Y^{(t)}_{tr}$,
and $D^{(t)}_{te}$ indicates the testing set with $Y^{(t)}_{te}$.
The training and testing classes are disjoint, so that $Y^{(t)}_{tr} \cap Y^{(t)}_{te} = \O$. 
Note that, only $D^{(t)}_{tr}$ is available at the $t$-th training step, and the data from previous domains are
\textit{not} available any more.
For evaluation, we test retrieval performance on all encountered domains with their corresponding testing sets.
In addition, the generalization ability is evaluated via new and unseen domains $D_{un}$ with unseen identities $Y_{un}$.
Henceforth, we will drop the subscript $\{tr, te\}$ for simplicity of notation. 

\begin{figure*}[th]
  \begin{center}
  \includegraphics[width=170mm]{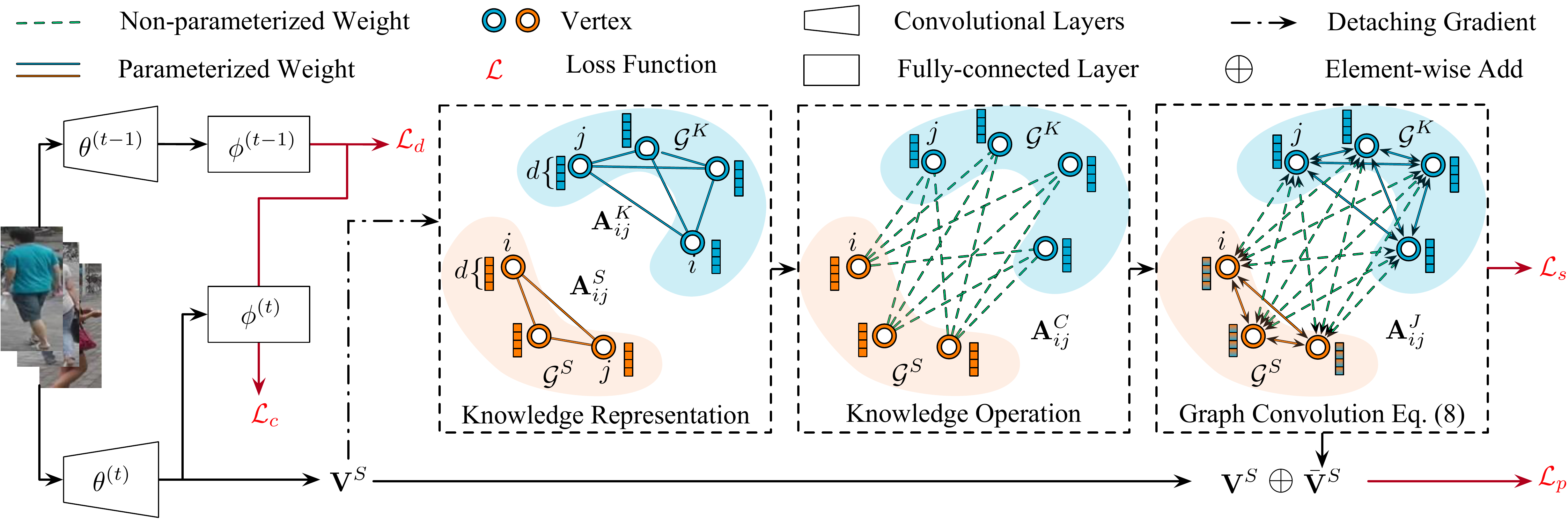}
  \captionsetup{aboveskip=0pt}
\caption{Overview of the proposed AKA framework. AKA maintains the AKG parameterized by $\psi$, to organize and memorize previous learned knowledge. Given a mini-batch images from a certain domain, similarity graph $\mathcal{G}^{S}$ is constructed by the extracted features $\mathbf{V}^{S}$. Meanwhile it taps into AKA to acquire relevant knowledge from $\mathcal{G}^{K}$, resulting in the vectored representations $\bar{\mathbf{V}}^{S}$ of acquired knowledge. Further, the required knowledge $\bar{\mathbf{V}}^{S}$ are summed with corresponding input features $\mathbf{V}^{S}$, which generates enhanced representation with better generalization capability. \label{fig:graph}}
  \end{center}
  \vspace{-0.5cm}
\end{figure*}


\subsection{Baseline Approach}\label{sec:baseline}
We introduce a baseline solution based on knowledge distillation to address LReID. The baseline model consists of a feature extractor $h\left(\cdot;\theta\right)$ 
with parameters $\theta$ and a classifier $g\left(\cdot;\phi\right)$ with parameters $\phi$. The whole network $f\left(\cdot;\theta,\phi\right)$ is the mapping from the input space directly to confidence scores, which is defined as: $f\left(\cdot;\theta,\phi\right):= g\left(h\left(\cdot;\theta\right);\phi\right)$. Training the parameters $\theta$ and $\phi$ in the network is optimized by a cross-entropy loss,
\begin{equation}
\mathcal{L}_{c}=-\sum_{(\mathbf{x},\mathbf{y})\in D} \mathbf{y} \log \left(\bm{\sigma}\left(f\left(\mathbf{x};\theta,\phi\right)\right)\right),
\end{equation}
where $\bm{\sigma}$ is \textit{softmax} function.
In addition, we adopt the knowledge distillation (KD) \cite{li2017learning} 
technique for mitigating forgetting on previous domains. Omitting the superscript $(t)$, the loss function is defined as:
\begin{equation}
\begin{split}
    \mathcal{L}_{d}&=\\
    &-\sum_{\mathbf{x}\in D}\sum_{j=1}^{n}\bm{\sigma}\left(f(\mathbf{x}; \hat{\theta},\hat{\phi})\right)_{j} \log \left(\bm{\sigma}\left(f(\mathbf{x};\theta,\phi)\right)_{j}\right),
\end{split}
\end{equation}
where $n=\sum^{t-1}_{i=1}\left|Y^{(i)}\right|$ is the number of the old classes, $\hat{\theta}$ and $\hat{\phi}$ are copied from $\theta$ and $\phi$ before current-step training, respectively. The total objective of baseline method is:
\begin{equation}\label{eq:base}
    \mathcal{L}_{base} =\mathcal{L}_{c} + \gamma\mathcal{L}_{d},
\end{equation}
where $\gamma$ is a trade-off factor for the knowledge distillation loss and the cross-entropy loss.

\section{Adaptive Knowledge Accumulation}
In this section, we introduce the details of the proposed AKA framework. The goal of AKA is to facilitate both learning process of new domain and generalization on unseen domains by leveraging transferable knowledge learned from previous domains. Referring to biological prior knowledge, AKA mimics the brain's cognitive process \cite{cowell2019roadmap} to construct two sub-processes: \textit{knowledge representation} and \textit{knowledge operation}, illustrated by Fig.~\ref{fig:graph}. In the following subsections, we elaborate both sub-processes and their optimization, respectively.

\subsection{Knowledge Representation}\label{sec:similarity_graph}
To respectively represent the knowledge underling current samples, and the accumulated knowledge learned from already-trained domains, we parameterize the knowledge \textit{``representations''} by constructing two different graph structures: \textit{instance-based similarity graph} (ISG) and \textit{accumulated knowledge graph} (AKG).


\noindent\textbf{Instance-based Similarity Graph.}
Given a mini-batch of samples from a certain domain, the extracted features are defined as $\mathbf{V}^{S}=h\left(\mathbf{x};\phi\right)$. Inspired by \cite{luo2019spectral}, we first investigate the relationships among these samples and represent the relationships by a fully-connected graph $\mathcal{G}^{S}(\mathbf{A}^{S}, \mathbf{V}^{S})$, namely ISG, where $\mathbf{A}^{S}$ is the edge set and the extracted features serve as vertices $\mathbf{V}^{S}$ in the graph. 
The edge weight $\mathbf{A}^{S}_{ij}$ between two vertices $\mathbf{V}^{S}_{i}$ and $\mathbf{V}^{S}_{j}$ is measured by a learnable $L_{1}$ distance between them:
\begin{equation}\label{eq:similarity_adj}
    \mathbf{A}^{S}_{ij} = \bm{\rho} \big(\mathbf{W}^{S}\left | \mathbf{V}^{S}_{i} - \mathbf{V}^{S}_{j} \right | + \mathbf{b}^{S} \big),
\end{equation}
where $\mathbf{W}^{S}$ and $\mathbf{b}^{S}$ represent learnable parameters, and $\bm{\rho}$ is \textit{Sigmoid} function. This is, the ISG is build by parameterized weight as shown in Fig.~\ref{fig:graph}. For each mini-batch with $N^{b}$ samples, AKA temporarily constructs a $\mathcal{G}^{S}$, in which $\mathbf{V}^{S} \in \mathbb{R}^{N^{b}\times d}$ denotes a feature set with dimensions $d$ and $\mathbf{A}^{S} \in \mathbb{R}^{N^{b}\times N^{b}}$ gives the adjacency matrix. 
This matrix indicates the proximity between instances.
\par
\noindent\textbf{Accumulated Knowledge Graph.}
Furthermore, to represent accumulated knowledge, we construct an AKG, whose vertices represent different types of knowledge (e.g., the representative person appearance and structure) and edges are automatically constructed to reflect the relationship between such knowledge. Specifically,
Given an vertex set $\mathbf{V}^{K} \in \mathbb{R}^{N^{k}\times d}$ and an adjacent matrix $\mathbf{A}^{K} \in \mathbb{R}^{N^{k}\times N^{k}}$, we define the knowledge graph as $\mathcal{G}^{K}(\mathbf{A}^{K}, \mathbf{V}^{K})$, where $N^{k}$ is the number of the AKG's vertices. To better explain the construction of the AKG, we first discuss the vertex representation $\mathbf{V}^{K}$.
During domain-incremental training, domains arrive 
sequentially and their corresponding vertices representations are expected to be updated dynamically and timely. 
Therefore, the vertex representations of the AKG is parameterized and learned at the training time. Moreover, to encourage the diversity of knowledge encoded in the AKG, the vertex representations are randomly initialized. Analogous to the definition of weight in the ISG, the parameterized weight of AKG 
is defined as:
\begin{equation}\label{eq:knowledge_adj}
    \mathbf{A}^{K}_{ij} = \bm{\rho}\big(\mathbf{W}^{K}(\left | \mathbf{V}^{K}_{i} - \mathbf{V}^{K}_{j} \right |) + \mathbf{b}^{K} \big),
\end{equation}
where $\mathbf{W}^{K}$ and $\mathbf{b}^{K}$ represent learnable parameters. 
\par
\noindent\textbf{Remark:} The weights in $\mathcal{G}^{S}$ and $\mathcal{G}^{K}$ are calculated by independent learnable parameters, as the manners of knowledge organization in two graph have distinct differences. One focuses on the relationship among current samples. The other is required to consider both its own structure and efficient knowledge transformation. Such design is distinct different from the graph matching network \cite{li2019graph} that shares same weights of two graphs like a Siamese network. 

\subsection{Knowledge Operation}
Based on such knowledge representations, we further decompose the \textit{``operations''} into \textit{knowledge transfer} and \textit{knowledge accumulation}, to enhance the learning of new domains with involvement of previous knowledge, and update these accumulated knowledge, correspondingly.

\noindent\textbf{Knowledge Transfer.} We first discuss how to organize and extract knowledge from the previous learning process and then explain how to leverage such knowledge to benefit the training of a new domain. The edges in $\mathcal{G}^{S}$ and $\mathcal{G}^{K}$ are also reserved in the joint graph $\mathcal{G}^{J}$. We connect $\mathcal{G}^{S}$ with $\mathcal{G}^{K}$ by creating links between the prototype-based relational graph 
and the knowledge graph. The cross-graph edge between a pair of vertices in $\mathcal{G}^{S}$ and $\mathcal{G}^{K}$ is weighted by the similarity between them. Specifically, for 
each instance pair $\mathbf{V}^{S}_{i}$ and $\mathbf{V}^{K}_{j}$, the cross-graph weight $\mathbf{A}^{C}_{ij}$ is calculated by applying a Softmax over Euclidean distances between $\mathbf{V}^{S}_{i}$ and $\mathbf{V}^{K}_{j}$, which is a non-parameterized similarity:
\begin{equation}\label{eq:joint_adj}
    \mathbf{A}^{C}_{ij}=\frac{\exp (-\frac{1}{2}\left \| \mathbf{V}^{S}_{i} - \mathbf{V}^{K}_{j} \right \|^{2}_{2})}{\sum^{N^{k}}_{k=1}\exp (-\frac{1}{2}\left \| \mathbf{V}^{S}_{i} - \mathbf{V}^{K}_{k} \right \|^{2}_{2})}.
\end{equation}
Taking Eq. \ref{eq:similarity_adj}, \ref{eq:knowledge_adj} and \ref{eq:joint_adj}, the joint graph is formulated as:
\begin{equation}
    \mathbf{A}^{J} =\begin{bmatrix}
\mathbf{A}^{S} & \mathbf{A}^{C} \\ 
 (\mathbf{A}^{C})^{T}& \mathbf{A}^{K}
\end{bmatrix},     \mathbf{V}^{J} =\begin{bmatrix}
\mathbf{V}^{S} \\ 
 \mathbf{V}^{K}
\end{bmatrix},
\end{equation}
where the adjacent matrix $\mathbf{A}^{J} \in \mathbf{R}^{(N^{b} + N^{k})\times (N^{b} + N^{k})}$ and vertex matrix $\mathbf{V}^{J} \in \mathbf{R}^{(N^{b} + N^{k})\times d}$ define joint graph $\mathcal{G}^{J}$.
\par
After constructing the joint graph $\mathcal{G}^{J}$, we 
propagate the most related knowledge from $\mathcal{G}^{K}$ to $\mathcal{G}^{S}$ via a Graph Convolutional Network (GCN) \cite{kipf2016semi}, which is formulated as:
\begin{equation}
\mathbf{V}^{G} = \bm{\delta} \big(\mathbf{A}^{J}(\mathbf{V}^{J}\mathbf{W}^{J})\big),
\end{equation}
where $\mathbf{V}^{G} \in \mathbf{R}^{(N^{b} + N^{k})\times d}$ is the vertex embedding after one-layer ``message-passing'' \cite{gilmer2017neural} and $\mathbf{W}^{J}$ is a learnable weight matrix of the GCN layer followed by a non-linear function $\bm{\delta}$, e.g., ReLU \cite{agarap2018deep}. We employ only one layer to accomplish information propagation for simplicity, while it is natural to stack more GCN layers. After passing features through GCN, we obtain the information-propagated feature representation of the $\mathbf{V}^{S}$ from the top-$N^{b}$ rows of $\mathbf{V}^{G}$, which is denoted as $ \bar{\mathbf{V}}^{S}= \{\mathbf{V}^{G}_{i} |i \in [1, N^{b}]\}$.
\par
\noindent\textbf{Knowledge Accumulation.} Maintaining a knowledge graph within limited storage resource during lifelong learning is inevitably expected to compact memorized knowledge and selectively update the AKG. To achieve 
this goal, we first aggregate $\mathbf{V}^{S}$ and $\bar{\mathbf{V}}^{S}$ by summing them, which results in a set of summed representation $\mathbf{F} = \left(\mathbf{V}^{S} +\bar{\mathbf{V}}^{S}\right)/2$. Then, to guide $\bar{\mathbf{V}}^{S}$ that improves the generalization of $\mathbf{V}^{S}$, we introduce a plasticity objective:
\begin{equation}\label{eq:plasticity}
\mathcal{L}_{p}= \frac{1}{N^{b}}\sum_{(a,p,n)}\ln\Big(1+\exp{\big(\Delta(\mathbf{F}_{a}, \mathbf{F}_{p})-\Delta(\mathbf{F}_{a}, \mathbf{F}_{n})\big)}\Big),\\
\end{equation}
where $\Delta$ denotes a distance function, e.g., $L_{2}$ distance and cosine distance. $a, p$ and $n$ donate the anchor, positive and negative instances in a mini-batch while we utilize an online hard-mining sampling strategy \cite{ye2020augmentation} to boost generalization capability of learned representation. 
\par
Furthermore, we observed that only encouraging the knowledge graph to adapt the current domain easily results in significant over-fitting, which would further lead to catastrophic forgetting. Thus, we propose a stability loss to punish the large movements of vertices in $\mathcal{G}^{K}$ when they update from the ending state $\hat{\mathbf{V}}^{K}$ of last training step:
\begin{equation}\label{eq:stability}
\mathcal{L}_{s}=\frac{1}{N^{k}}\sum_{i=1}^{N^{k}}\ln\Big(1+\exp{\big(\Delta(\mathbf{V}^{K}_{i}, \hat{\mathbf{V}}^{K}_{i}\big)}\Big).
\end{equation}
This loss term constrains the vertices in $\mathcal{G}^{K}$ to approximate their initial parameters. Eq.~\ref{eq:plasticity} and Eq.~\ref{eq:stability} are used to co-optimize the parameters of AKG but detaching the gradient flowing into CNN, which is discussed in Sec.~\ref{sec:discussion}. Through imposing such stability-plasticity dilemma, AKG accumulates more refine and general knowledge from comparison with previous knowledge, so as to generate better representation for generalizable ReID.

\subsection{Optimization}
According to \cite{cowell2019roadmap,wang2018knowledge}, when a visual cognitive process starts, our brain retrieves relevant representational content (knowledge) from high-dimensional memories based on similarity or familiarity. Then, our brain will summarize the captured information and update relevant knowledge or allocate new memory. Motivated by this, we query the ISG in the AKG to obtain the relevant previous knowledge. The ideal query mechanism is expected to optimize both graphs simultaneously at the training time and guide the training of both graphs to be mutual promotion. At the training step $t$, we train the whole model $\Theta^{(t)}=\{\theta^{(t)},\phi^{(t)},\psi^{(t)}\}$ on $D^{(t)}$ with mini-batch SGD and detaching the gradient between $\theta^{(t)}$ and $\psi^{t}$. The overall loss function is:
\begin{equation}
\begin{split}
\mathcal{L}_{total} = \mathcal{L}_{base} +  \lambda_{p}\mathcal{L}_{p} + \lambda_{s}\mathcal{L}_{s},\\
\end{split}
\end{equation}
where $\lambda_{s}$ and $\lambda_{p}$ are plasticity-stability trade-off factors. Here, we discuss how our proposed AKG works. When $\lambda_{p}$ is relatively larger than $\lambda_{s}$, $\mathcal{G}^{K}$ focuses on learning new knowledge with minimal weight on taking into account previous knowledge.
On the contrary, our model can only benefits for improving generalization in first two domain-incremental steps with approximately fixed vertices of knowledge graph. 
Intuitively, the optimal balance of these two terms not only ensures the stability of knowledge graph, but also endows AKG with a plasticity that allows new knowledge to be incorporated and accumulated.

\subsection{Discussion}\label{sec:discussion}
\noindent\textit{(1) Why does AKA respectively use non-parameterized and parameterized weight for knowledge operation and representation?}
In the sight of \cite{kirkpatrick2017overcoming}, the partial parameters of top layers favor becoming domain-specific during incremental training on different domains, which leads to severe performance degradation on previous domains. In addition, according to the biological inspiration \cite{cowell2019roadmap}, the representation and operation should be independent. To this end, when performing knowledge transformation, a non-parameterized metric allows model to treat different domains with less bias. As for the knowledge representation, summarizing and updating knowledge require the power of parameters.

\noindent\textit{(2) Why does AKA detach the gradient of GCN?} As shown in Fig.~\ref{fig:ab}, AKA without detaching gradient tends to transfer relatively similar knowledge through all training domains, which is caused by the degradation of GCN~\cite{hospedales2020meta}. However, detaching the gradient encourages AKA to learn independently so that AKA enables to adaptively generate different knowledge for different domains. 
\par
\noindent\textit{(3) Why is the proposed straightforward $\mathcal{L}_{s}$ efficient?}
Intuitively, the unity of $\mathcal{L}_{s}$ and $\mathcal{L}_{p}$ forms a bottleneck mechanism, which forces $\mathcal{G}^{K}$ to learn sparse knowledge from each domain. In this work, we utilize a simple yet effective method, restricting the vertices only, to preserve knowledge. Even though the vertices are almost fixed, the weight of transferable knowledge is learnable. Ideally, $\mathcal{G}^{K}$ could adaptively modify the transformation weight so as to reorganize old knowledge for representing new knowledge. That means we maintain the topology of vertices and leverage flexible non-parameter transformation to adapt feature representations in a new environment. 

\begin{table*}[th]
\centering
\tiny
\captionsetup{aboveskip=0pt}
\caption{The statistics of ReID datasets involved in our experiments. `*' denotes that we modified the original dataset by using the ground-truth person bounding box annotation for our lifelong ReID experiments rather than using the original images which were originally used for person search evaluation. `-' denotes these data are not used for lifelong training.\label{tab:datasets}}
\resizebox{\textwidth}{!}{%
\begin{tabular}{c|l|l|lll|lll}
\hline
\multirow{2}{*}{Benchmark} & \multicolumn{1}{c|}{\multirow{2}{*}{Datasets Name}} & \multicolumn{1}{c|}{\multirow{2}{*}{Scale}} & \multicolumn{3}{c|}{Original Identities} & \multicolumn{3}{c}{Selected Identities} \\ \cline{4-9} 
 & \multicolumn{1}{c|}{} & \multicolumn{1}{c|}{} & \multicolumn{1}{c}{Train} & \multicolumn{1}{c}{Query} & \multicolumn{1}{c|}{Gallery} & \multicolumn{1}{c}{Train} & \multicolumn{1}{c}{Query} & \multicolumn{1}{c}{Gallery} \\ \hline
\multirow{5}{*}{\begin{tabular}[c]{@{}c@{}}LReID-Seen\end{tabular}} & CUHK03\cite{li2014deepreid} & mid & 767 & 700 & 700 & 500 & 700 & 700 \\
 & Market-1501\cite{zheng2015scalable} & large & 751 & 750 & 751 & 500 & 751 & 751 \\
 & MSMT17\_V2 \cite{wei2018person} & large & 1041 & 3060 & 3060 & 500 & 3060 & 3060 \\
 & DukeMTMC-ReID\cite{zheng2017unlabeled} & large & 702 & 702 & 1110 & 500 & 702 & 1110 \\
 & CUHK-SYSU ReID*\cite{xiao2016end} & mid & 942 & 2900 & 2900  & 500 & 2900 & 2900 \\ \hline
\multirow{7}{*}{\begin{tabular}[c]{@{}c@{}}LReID-Unseen\end{tabular}} & VIPeR\cite{gray2008viewpoint} & small & 316 & 316 & 316 & - & 316 & 316 \\
 & PRID\cite{hirzer2011person} & small & 100 & 100 & 649 & - & 100 & 649 \\
 & GRID\cite{loy2010time} & small & 125 & 125 & 126 & - & 125 & 126 \\
 & i-LIDS\cite{wei2009associating} & small & 243 & 60 & 60 & - & 60 & 60 \\
 & CUHK01\cite{li2012human} & small & 485 & 486 & 486 & - & 486 & 486 \\
 & CUHK02\cite{li2013locally} & mid & 1677 & 239 & 239 & - & 239 & 239 \\
 & SenseReID\cite{zhao2017spindle} & mid & 1718 & 521 & 1718 & - & 521 & 1718 \\ \hline
\end{tabular}%
}
\vspace{-0.5cm}
\end{table*}

\section{Experiments}
\subsection{Implementation Details}
We remove the last classification layer of ResNet-50 and use the retained layers as the feature extractor to yield 2048-dimensional features. The AKA network consists of one GCN layer. In each training batch, we randomly select 32 identities and sample 4 images for each identity. All images are resized to 256 $\times$ 128. Adam optimizer with learning rate $3.5\times10^{-4}$ is used. The model is trained for 50 epochs, and decrease the learning rate by $\times$ 0.1 at the 25$^{th}$ and 35$^{th}$ epoch. We follow \cite{zhao2020continual} to set the balance weight $\gamma$ as 1, and explore the effect of other hyper-parameters. The $N^{K}$, $\lambda_{p}$, and $\lambda_{s}$ are set as 64, 1, and 10, respectively. The hyper-parameter analysis is given in Sec.~\ref{sec:ablation_study}. The retrieval of testing data is based on Euclidean distance of feature embeddings. For all experiments, we repeat five times and report means and standard deviations. 
\subsection{New Benchmark for LReID}
\par
We present a new and large-scale benchmark including LReID-Seen and LReID-Unseen subsets. The presented benchmarks are different from existing ReID benchmarks in three main aspects: 1) The proposed LReID benchmarks are specifically designed for person re-identification that is the fine-grained retrieval task, while existing lifelong learning benchmarks focus on general image classification; 2) The total number of classes in our benchmark ($\left|Y\right|\approx$14K) is much larger than existing benchmarks ($\leq$ 1K); 3) We test the model on novel identities that have never appeared in the training set even on unseen domains, while existing benchmarks test on new images of learned (seen) classes. 

\begin{table*}[ht]
\centering
\captionsetup{aboveskip=0pt}
\caption{Seen-domain non-forgetting evaluation. We test model after sequentially training on all seen domains ($t$=5). Each experiment is repeated by 5 times to report mean and std of all seen domains. The training order is MA$\rightarrow$SY$\rightarrow$DU$\rightarrow$MS$\rightarrow$CU.\label{tab:forget}}
\resizebox{\textwidth}{!}{%
\begin{tabular}{c|cc|cc|cc|cc|cc|cc}
\cline{1-13}
 & \multicolumn{2}{c|}{Market} & \multicolumn{2}{c|}{SYSU} & \multicolumn{2}{c|}{Duke} & \multicolumn{2}{c|}{MSMT17} & \multicolumn{2}{c|}{CUHK03} & \multicolumn{2}{c}{$\bar{s}$}  \\ \cline{2-13}
Method & mAP & Rank-1 & mAP & Rank-1 & mAP & Rank-1 & mAP & Rank-1 & mAP & Rank-1 & mAP & Rank-1 \\ \cline{1-13}
SFT & 24.1$\pm$0.2 & 48.5$\pm$0.4 & 30.5$\pm$0.3 & 32.7$\pm$0.3 & 14.4$\pm$0.2 & 27.0$\pm$0.2 & 12.0$\pm$0.3 & 30.1$\pm$0.3 & 45.6$\pm$0.2 & 48.5$\pm$0.3 & 25.3 & 37.4 \\
SFT-T & 25.8$\pm$0.3 & 48.4$\pm$0.5 & 32.0$\pm$0.4 & 34.8$\pm$0.5 & 15.1$\pm$0.4 & 27.7$\pm$0.5 & 13.8$\pm$0.3 & 32.4$\pm$0.4 & \textbf{48.0}$\pm$0.4 & \textbf{50.1}$\pm$0.5 & 26.9 & 38.7 \\
SPD & 30.5$\pm$0.3 & 50.7$\pm$0.4 & 37.6$\pm$0.4 & 39.9$\pm$0.4 & 14.6$\pm$0.3 & 27.2$\pm$0.3 & 12.2$\pm$0.3 & 30.3$\pm$0.2 & 40.7$\pm$0.3 & 42.5$\pm$0.3 & 27.1 & 38.1\\
LwF & 47.1$\pm$0.3 & 65.1$\pm$0.4 & 43.9$\pm$0.2 & 44.3$\pm$0.2 & 16.0$\pm$0.3 & 28.0$\pm$0.2 & 14.8$\pm$0.2 & 33.6$\pm$0.3 & 26.1$\pm$0.2 & 26.2$\pm$0.2 & 29.6 & 39.4\\
CRL & 48.5$\pm$0.5 & 66.6$\pm$0.3 & 45.2$\pm$0.2 & 43.3$\pm$0.3 & 16.2$\pm$0.2 & 27.9$\pm$0.3 & 16.1$\pm$0.3 & 34.3$\pm$0.2 & 28.1$\pm$0.3 & 29.8$\pm$0.4 & 30.8 & 40.4 \\
CRL-T & 49.2$\pm$0.5 & 67.0$\pm$0.3 & 45.6$\pm$0.2 & 43.9$\pm$0.4 & 15.9$\pm$0.3 & 27.5$\pm$0.4 & 15.8$\pm$0.3 & 33.9$\pm$0.4 & 26.5$\pm$0.3 & 26.7 $\pm$0.4 & 30.6 & 39.8 \\
AKA & \textbf{51.2}$\pm$0.2 & \textbf{72.0}$\pm$0.3 & \textbf{47.5}$\pm$0.5 & \textbf{45.1}$\pm$0.6 & \textbf{18.7}$\pm$0.3 & \textbf{33.1}$\pm$0.4 & \textbf{16.4}$\pm$0.2 & \textbf{37.6}$\pm$0.3 & 27.7$\pm$0.4 & 27.6$\pm$0.5 & \textbf{32.3} & \textbf{43.1} \\ \cline{1-13}
Joint-CE & 71.9$\pm$0.2 & 83.2$\pm$0.2 & 61.2$\pm$0.3 & 62.5$\pm$0.3 & 65.1$\pm$0.2 & 76.8$\pm$0.3 & 25.3$\pm$0.3 & 50.7$\pm$0.5 & 48.7$\pm$0.1 & 50.3$\pm$0.2 & 54.4 & 64.7 \\
Joint-CE-T & 74.8$\pm$0.2 & 87.0$\pm$0.3 & 63.3$\pm$0.3 & 65.5$\pm$0.4 & 68.3$\pm$0.2 & 80.1$\pm$0.3 & 27.9$\pm$0.3 & 54.1$\pm$0.5 & 50.8$\pm$0.2 & 56.6$\pm$0.2 & 57.0 & 67.7  \\ \cline{1-13}
\end{tabular}%
}

\end{table*}


\begin{figure*}
     \centering
\includegraphics[width=\textwidth]{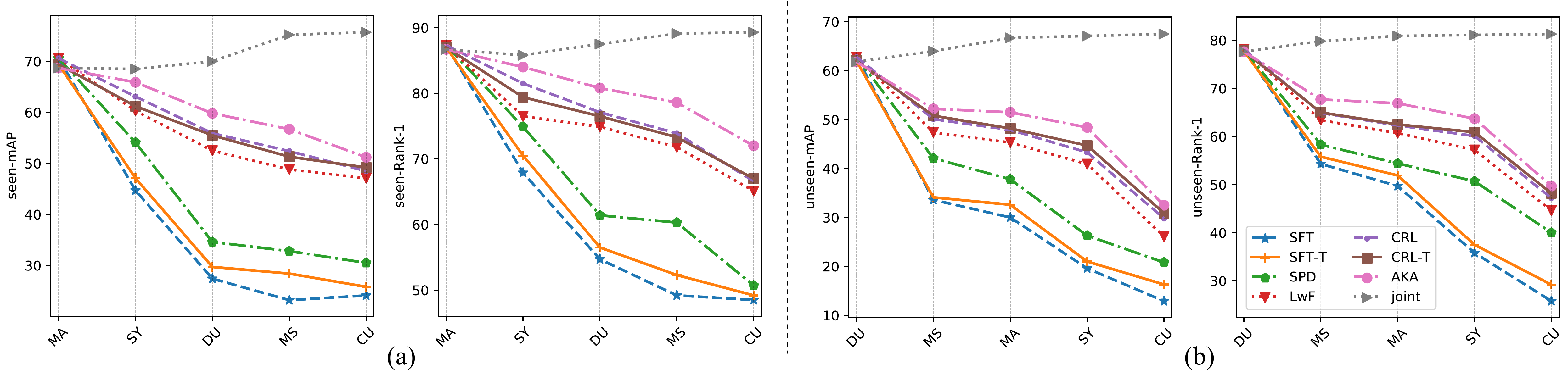}
     \captionsetup{aboveskip=0pt}
        \caption{Illustration of seen-domain non-forgetting evaluation. (a) depicts the trend of mAP and Rank-1 score on the first training domain during training process following \textit{Order-1}. 
        Likewise, (b) shows the results of \textit{Order-2}.\label{fig:forgetting}} 
        \vspace{-0.5cm}
\end{figure*}

     



\par
\noindent
\textbf{LReID-Seen.} In total, 40,459 training images of the 2,500 identities are employed for the lifelong training set. The training identities are uniformly split into 5 subsets in accordance with their domains, for 5-step domain-incremental training. Their original testing sets are kept to evaluate the model's domain forgetting and performance of the current domain. Specifically, we selected five relatively large-scale datasets, CUHK03~(CU)~\cite{li2014deepreid}, Market-1501~(MA)~\cite{zheng2015scalable}, MSMT17\_V2~ (MS)~\cite{wei2018person}, DukeMTMC-ReID~(DU)~\cite{zheng2017unlabeled} and CUHK-SYSU ReID~(SY)~\cite{xiao2016end}, and sampled 500 identities from each of their training sets to construct five training domains so that each domain has an equal number of classes. Note that for the SY~\cite{xiao2016end} dataset, we modified the original dataset by using the ground-truth person bounding box annotation and selected a subset in which each identity includes at least 4 bounding boxes, rather than using the original images which were originally used for person search evaluation. For testing on this dataset, we fixed both query and gallery sets instead of using variable gallery sets. We used 2,900 query persons, with each query containing at least one image in the gallery, which resulted in 942 training identities, called CUHK-SYSU ReID in Tab.~\ref{tab:datasets}. 
\par
\noindent\textbf{LReID-Unseen.} To verify raising the model's abilities resulting from progressively accumulated knowledge from previous domains, we reorganize 7 popular person ReID datasets as shown in Tab.~\ref{tab:datasets}. Specifically, we first merge VIPeR~\cite{gray2008viewpoint}, PRID~\cite{hirzer2011person}, GRID~\cite{loy2010time}, i-LIDS~\cite{wei2009associating}, CUHK01~\cite{li2012human}, CUHK02~\cite{li2013locally}, SenseReID~\cite{zhao2017spindle} in accordance with their original train/test splits as a new benchmark. Then, the merged test set, including 3,594 different identities with total 9,854 images, is adopted to evaluate the generalization ability of learned features on unseen domain, called LReID-Unseen in Tab.~\ref{tab:datasets}.

\par
\noindent\textbf{Evaluation metrics.}
We use $\bar{u}$ (average performance on unseen domains) to measure the capacity of generalising on unseen domains and $\bar{s}$ (average performance on seen domains) to measure the capacity of retrieving incremental seen domains. Note that the performance gap of $\bar{s}$ between joint training and a certain method indicates the method's ability to prevent forgetting. $\bar{u}$ and $\bar{s}$ are measured with mean average precision (mAP) and rank-1 (R-1) accuracy. These metrics are calculated after the last training step.

\begin{table*}[ht]
\centering
\captionsetup{aboveskip=0pt}
\caption{Unseen-domain generalising evaluation. We refer to corresponding literature and reproduce experimental results on our setting.
For LReID-Unseen, the training order is MA$\rightarrow$SY$\rightarrow$DU$\rightarrow$MS$\rightarrow$CU. \label{tab:generalising} }
\resizebox{\textwidth}{!}{%
\begin{tabular}{c|c|ccccccc|cc}
\hline
Banchmark & $\bar{u}$ & SFT & SFT-T & SPD & LwF & CRL & CRL-T & AKA & Joint-CE & Joint-CE-T \\ \hline
\multirow{2}{*}{\begin{tabular}[c]{@{}c@{}}CRL-ReID\\ (5-step)\end{tabular}} & mAP & 44.2 $\pm$ 0.2& 44.7 $\pm$ 0.3 & 47.1 $\pm$ 0.2 & 48.7 $\pm$ 0.2 & 51.2 $\pm$ 0.1 & 51.5 $\pm$ 0.2 & \textbf{64.2} $\pm$ 0.1 & 64.8 $\pm$ 0.1 & 66.7 $\pm$ 0.1 \\ \cline{2-11} 
 &  R-1 & 53.4 $\pm$ 0.3 & 53.9 $\pm$ 0.4 & 54.1 $\pm$ 0.4 & 59.6 $\pm$ 0.2 & 62.8 $\pm$ 0.3 & 63.1 $\pm$ 0.3 & \textbf{74.9} $\pm$ 0.3 & 75.3 $\pm$ 0.1 & 78.6 $\pm$ 0.2 \\  \hline
\multirow{2}{*}{\begin{tabular}[c]{@{}c@{}}CRL-ReID\\ (10-step)\end{tabular}} & mAP  & 31.7 $\pm$ 0.2 & 31.7 $\pm$ 0.3 &40.3 $\pm$ 0.3 & 42.8 $\pm$ 0.2 & 43.8 $\pm$ 0.3 &  44.1 $\pm$ 0.1 & \textbf{49.7} $\pm$ 0.2 & 64.8 $\pm$ 0.1 & 66.7 $\pm$ 0.1 \\ \cline{2-11} 
 & Rank-1 & 40.3 $\pm$ 0.4 & 40.5 $\pm$ 0.5 & 47.5 $\pm$ 0.4 & 51.7 $\pm$ 0.1 & 54.7 $\pm$ 0.4 & 54.8 $\pm$ 0.3 & \textbf{58.8} $\pm$ 0.2 & 75.3 $\pm$ 0.1 & 78.6 $\pm$ 0.2 \\ \hline
\multirow{2}{*}{LReID-Unseen} & mAP &35.2$\pm$ 0.2&37.1$\pm$ 0.4&36.3$\pm$ 0.2&38.3$\pm$ 0.2&38.5$\pm$ 0.2&39.6$\pm$ 0.4&\textbf{44.3}$\pm$ 0.2&50.6$\pm$ 0.1  &53.5$\pm$ 0.2 \\ \cline{2-11} 
 & R-1 & 31.1$\pm$ 0.3&34.3$\pm$ 0.4& 32.9$\pm$ 0.2&36.9$\pm$ 0.3&36.7$\pm$ 0.2&38.1$\pm$ 0.4&\textbf{40.4}$\pm$ 0.3 &48.1 $\pm$ 0.1& 50.0 $\pm$ 0.3  \\ \hline
\end{tabular}
}  
\end{table*}


     

\begin{figure*}
     \centering
\includegraphics[width=\textwidth]{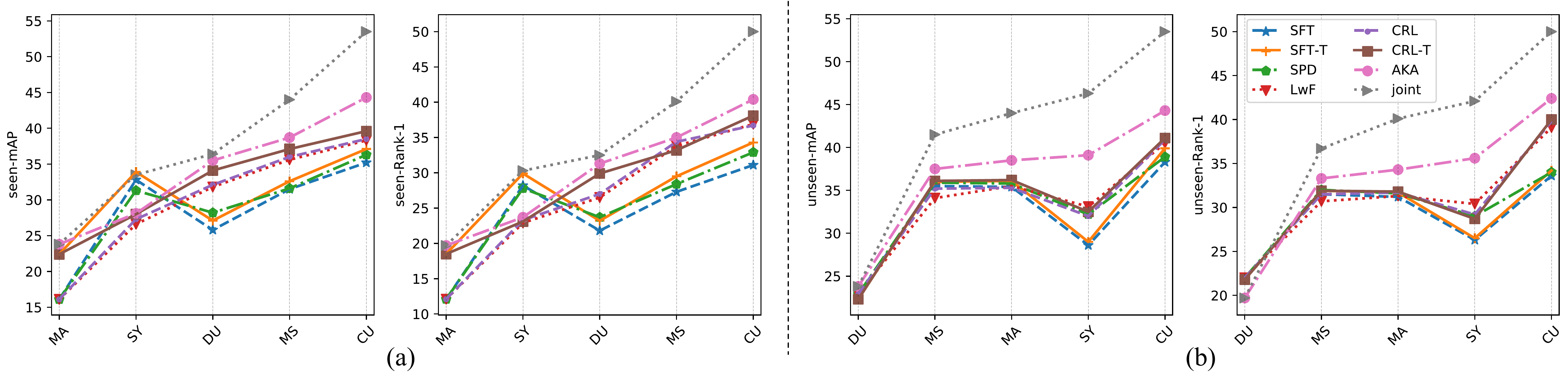}
     \captionsetup{aboveskip=0pt}
        \caption{Illustration of unseen-domain generalising evaluation. (a) depicts the trend of mAP and Rank-1 score on unseen domains during training process following \textit{Order-1}. Likewise, (b) shows the results of \textit{Order-2}.\label{fig:generalising}} 
        \vspace{-0.5cm}
\end{figure*}


\subsection{Seen-domain Non-forgetting Evaluation}
Less forgetting performance refers to the effectiveness of one method which mitigates the accuracy degradation on previous domains. We evaluated AKA on LReID task against the state-of-the-art. The methods for comparison include 1) sequential fine-tuning (SFT): Fine-tuning model with new datasets without distilling old knowledge; 2) learning without forgetting (LwF): The baseline method \cite{li2017learning} introduced in Sec.~\ref{sec:baseline}; 3) similarity-preserving distillation (SPD): A competitor with advanced feature distillation \cite{tung2019similarity}; 4) Continual representation learning (CRL) \cite{zhao2020continual}: We first reproduce their method and achieve the reported results on their published benchmark. Then, we apply their methods to our domain-incremental person ReID benchmark and report these new results in Table.~\ref{tab:forget}; 5) Joint-CE serves as an upper-bound by training model on all data of the seen domains with $\mathcal{L}_{c}$. For a fair comparison, SFT-T, CRL-T and Joint-CE denote directly adding the widely-used triplet loss \cite{hermans2017defense} for co-optimizing learned features. 
\par
In practice, the order of input domains is agnostic. Thus, we investigate the influence caused by different training orders and analyze two representative results. 
\textit{Order-1} and \textit{Order-2} are denoted by \textbf{MA}$\rightarrow$\textbf{SY}$\rightarrow$\textbf{DU}$\rightarrow$\textbf{MS}$\rightarrow$\textbf{CU} and \textbf{DU}$\rightarrow$\textbf{MS}$\rightarrow$\textbf{MA}$\rightarrow$\textbf{SY}$\rightarrow$\textbf{CU}, respectively. As shown in Fig.~\ref{fig:forgetting}, training order significantly impacts the model's ability to prevent forgetting. Specifically, for \textit{Order-1}, AKA ranks the first with accuracy degradation of 17.5\%/14.7\% in mAP/R-1, which demonstrates that AKA is able to preserve old knowledge while mitigating catastrophic forgetting. In comparison, AKA outperforms SFT by around 30\% in R-1 and is superior to most competitive CRL by 6\% in mAP. Note that SFT-T and CRL-T (with additional triplet loss) is not beneficial for the first three training steps, because when the number of training identities is large enough, triplet loss contributes less on performance and even leads to conflict with cross-entropy loss \cite{zheng2019pyramidal}. On the other hand, KD-based methods are obviously superior to feature distillation or SFT methods. For \textit{Order-2}, AKA ranks the first with performance degradation of 29.3\%/27.9\% in mAP/R-1 as well.

\subsection{Unseen-domain Generalising Evaluation}
To demonstrate that our LRe-ID is more challenging than the latest CRL-ReID~\cite{zhao2020continual} task, we re-implement their method and evaluate on both their CRL-ReID dataset~\cite{zhao2020continual} and our LReID-Unseen benchmarks. Despite our setting needs to overcome larger domain gaps, our AKA can automatically transfer and update knowledge based on different input. Thus, the results shown in the first two rows of Tab.~\ref{tab:generalising} indicate that LRe-ID setting is more difficult and our method outperforms the compared methods significantly.
\par
For the experiments on LReID-Unseen, we assumed that a model was sequentially trained with the \textit{Order-1}. Then, we report all results in the final step when all domains are trained. 
As shown in Tab.~\ref{tab:generalising}, AKA achieves best performance compared with other competitive methods. Specifically, AKA achieves averaged 31.8\% mAP on seen domains and averaged 44.3\% mAP on unseen domains, which are significantly better than the baseline methods. Interestingly, as shown in Fig.~\ref{fig:generalising}, the methods without KD reach a better performance on 2$^{nd}$ step, but they fail to accumulate previous knowledge to further improve generalization ability. The similar phenomenon appears in \textit{order-2} as well. However, our results are still obviously lower than the upper-bound. The gap indicates the challenges of LReID on the proposed benchmark.

\subsection{Ablation Study}\label{sec:ablation_study}

\begin{table}[ht]
\centering
\captionsetup{aboveskip=0pt}
\caption{Effectiveness of the proposed loss functions.\label{tab:ablation}}
\begin{tabular}{l|cc|cc}
\hline
\multicolumn{1}{c|}{}  & \multicolumn{2}{c|}{$\bar{s}$} &\multicolumn{2}{c}{$\bar{u}$} \\ \cline{2-5} 
\multicolumn{1}{c|}{Setting} & mAP    & R-1   & mAP  & R-1  \\ \hline
Baseline                       &29.6    & 39.4& 38.3   & 36.9    \\

Baseline + $\mathcal{L}_{p}$   & 29.5   & 39.6& 41.6    &  38.3   \\

Baseline + $\mathcal{L}_{p}$ + $\mathcal{L}_{s}$ (Full)                      & \textbf{32.3}  & \textbf{43.1} &  \textbf{44.3}     &  \textbf{40.4}  \\ \hline
Full w/o $\mathcal{L}_{d}$ &28.5 &39.1 &42.1&38.9\\ \hline
\end{tabular}
\vspace{-1.5em}
\end{table}

We conduct two groups of ablation experiments to study the effectiveness of our method. One is to verify the improvement of adding the AKG module. Our full method AKA is composed of LwF and AKG. Comparing the performances of LwF and AKA in Tab.~\ref{tab:forget}, our AKA achieves 6\% improvement on both mAP and less forgetting score. The other group is to demonstrate the importance of our proposed stability and plasticity loss. In Tab.~\ref{tab:ablation}, ``Baseline'' setting is the same as the LwF method. ``Baseline + $\mathcal{L}_{p}$'' denotes LwF method added our AKG with only plasticity loss. The ``Baseline + $\mathcal{L}_{p}$ + $\mathcal{L}_{s}$ '' setting indicates our full method. As shown in Tab.~\ref{tab:ablation}, $\mathcal{L}_{p}$ is beneficial for only unseen domains, and $\mathcal{L}_{p}$ and $\mathcal{L}_{s}$ are complementary. The improvement of adding $\mathcal{L}_{s}$ indicates that greater stability of knowledge can preserve the knowledge of previous domains, which remits the unfavourable influence of catastrophic forgetting to some extent. Moreover, the improvement of adding $\mathcal{L}_{p}$ indicates AKG is encouraged to learn how to transfer positive knowledge to improve generalization. When $\lambda_{p}$ becomes large enough, the model overfits on generating the same representation with the output of CNN.

\par
\begin{figure}[th]
  \begin{center}
  \includegraphics[width=0.48\textwidth]{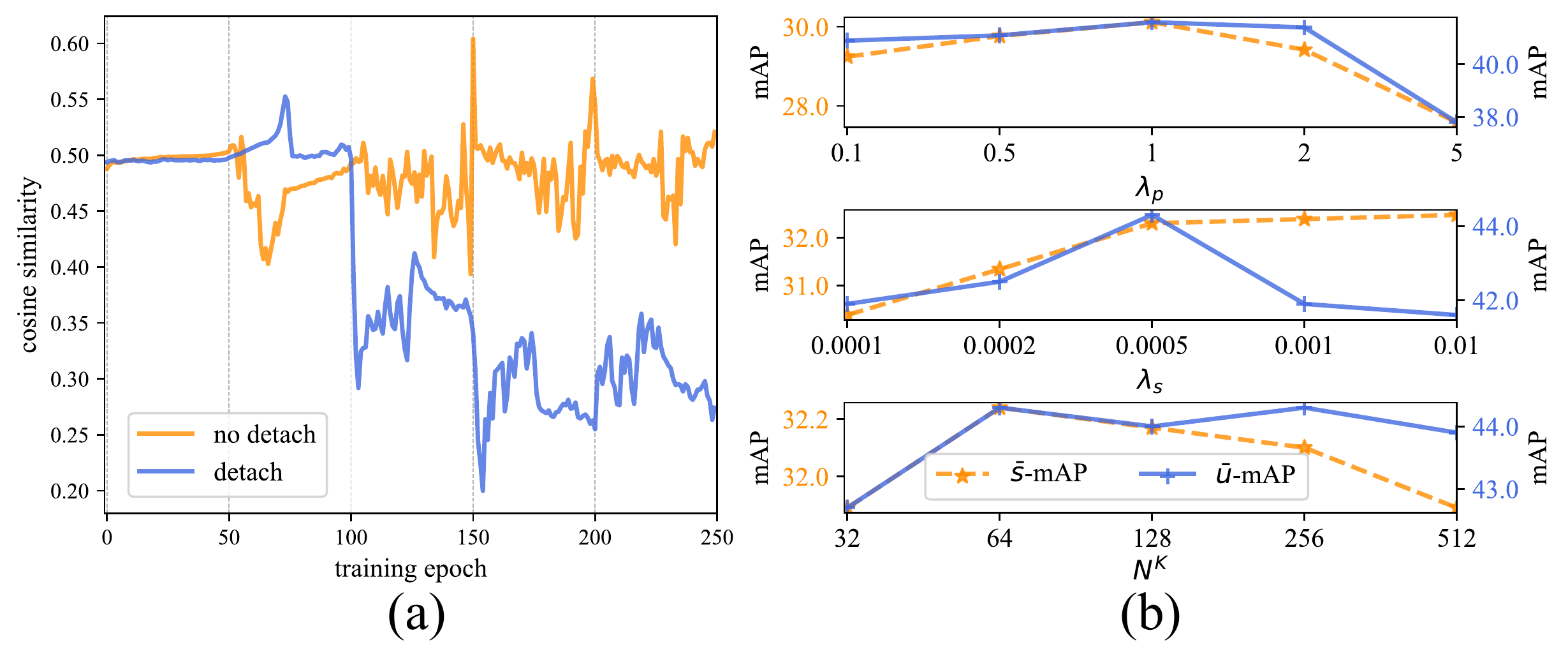}
  \captionsetup{aboveskip=0pt}
\caption{To investigate the effectiveness of detaching gradient, we visualize the normalized cosine similarity between $\mathbf{V}^{S}$ and $\bar{\mathbf{V}}^{S}$ during training processing in (a). The three rows in (b) study the effects of hyper-parameters $\lambda_{p}$, $\lambda_{s}$ and $N^{K}$, respectively. \label{fig:ab}}
  \end{center}
  \vspace{-0.1cm}
\end{figure}

\noindent\textbf{Hyper-parameter analysis.} The hold-off validation data are used to determine two hyper-parameters $\lambda_{p}$ and $\lambda_{s}$. We first select the optimal $\lambda_{p}$ to achieve best $\bar{u}$, then we choose the optimal $\lambda_{s}$ based on the selected $\lambda_{p}$. Finally, when $\lambda_{p}=1$ and $\lambda_{s}=5\times10^{-4}$, our model achieves best balance between seen and unseen domains. Afterwards, we keep other hyper-parameters and explore the influence of $N^{K}\in \{32, 64, 128, 256,512\}$ for $\bar{u}$ and $\bar{s}$ metrics calculated by mAP. The results shown in Fig.~\ref{fig:ab} indicate that $N^{K}$ is not sensitive and $\bar{u}$ increases with the growth of $N^{K}$. Thus, we balance memory consumption and generalization performance, and set $N^{K}=64$ in all of our experiments.


\section{Conclusion}
We focus on an unsolved, challenging, yet practical domain-incremental scenario, namely lifelong person re-identification, where models are required to improve generalization capability on both seen and unseen domains by leveraging previous knowledge. Hence, we propose a new AKA framework to preserve the knowledge learned from previous domains while adaptively propagating the previously learned knowledge for improving learning on new domains. Extensive experiments show that our method outperforms other competitors 
in terms of both mitigating forgetting on seen domains and generalising on unseen domains.

\section*{Acknowledgements}
This work was supported mainly by the LIACS Media Lab at Leiden University and in part by the China Scholarship Council and the Fundamental Research Funds for the Central Universities. Finally, I would like to thank my wife Dr. Yuzhi Lai who gave me the invaluable love, care and encouragement in the past years.
\clearpage

{\small
\bibliographystyle{ieee_fullname}
\bibliography{string,egbib}
}

\end{document}